\documentclass[pmlr]{jmlr}

\usepackage{longtable}
\usepackage{mathabx}

\usepackage{booktabs}
\usepackage[load-configurations=version-1]{siunitx} 

\usepackage{multirow}
 
\theorembodyfont{\upshape}
\theoremheaderfont{\scshape}
\theorempostheader{:}
\theoremsep{\newline}

\jmlrvolume{1} 
\jmlryear{2019}
\jmlrworkshop{NeurIPS2019 Disentanglement Challenge}

\title[Improved Disentanglement through Aggregated Convolutional Feature Maps]{NeurIPS 2019 Disentanglement Challenge: Improved Disentanglement through Aggregated Convolutional Feature Maps}

  \author{\Name{Maximilian Seitzer} \Email{contact@max-seitzer.de}\\
  \addr Fraunhofer Institute for Integrated Circuits IIS, Intelligent Systems Group, Erlangen, Germany}

\begin{document}

\maketitle

\begin{abstract}
This report to our stage 1 submission to the NeurIPS 2019 disentanglement challenge presents a simple image preprocessing method for training VAEs leading to improved disentanglement compared to directly using the images.
In particular, we propose to use regionally aggregated feature maps extracted from CNNs pretrained on ImageNet.
Our method achieved the 2nd place in stage 1 of the challenge~\citep{AicrowdChallenge2019}.
Code is available at \url{https://github.com/mseitzer/neurips2019-disentanglement-challenge}.
\end{abstract}

\section{Introduction}
\label{sec:intro}

The representational power and utility of feature representations obtained from deep CNNs trained on large image datasets such as ImageNet \citep{Russakovsky2014ImageNetLS} is well-known.
Amongst others, they are routinely used by practitioners to improve performance in transfer learning scenarios, and form the basis for perceptual loss functions \citep{Johnson2016PerceptualLF}.
A common view explaining the success of deep convolutional representations is that they describe an image in an abstract, concise way, simplifying downstream tasks such as classification \citep{Bengio2012RepresentationLA}. 
Thus, a natural hypothesis to draw is that it is easier for a Variational Autoencoder (VAE) \citep{Kingma2013AutoEncodingVB} to disentangle the latent factors of variations from this abstract description than from the image itself.
Therefore, in our challenge submission, we employ pretrained CNNs to extract convolutional feature maps as a preprocessing step before training the VAE.
To reduce the high-dimensional feature maps and fit the challenge's resource restrictions, we propose to aggregate the feature maps using a regional pooling technique from the context of image retrieval.

\section{Method}

Our method consists of the following three steps: (1) from each image in the dataset, extract a convolutional feature map using a CNN pretrained on ImageNet (section \ref{subsec:extraction}), (2) each feature map is aggregated into a feature vector and stored in memory (section \ref{subsec:aggregation}), (3) a VAE is trained to reconstruct the feature vectors and disentangle the latent factors of variation (section \ref{subsec:training}).
Appendix \ref{apd:notes} contains further comments about the hyperparameter choices and lists some other approaches we tested for the challenge.

\subsection{Feature Map Extraction}
\label{subsec:extraction}

To extract convolutional feature maps from the images, we use the VGG19-BN\footnote{\url{https://download.pytorch.org/models/vgg19_bn-c79401a0.pth}} architecture \citep{Simonyan2014VeryDC} in the \texttt{torchvision} package. 
In particular, we use the pretrained weights stemming from training on ImageNet without further finetuning them in any way. 
Input images are transformed to the format the pretrained networks expect, i.\,e. we bilinearly resize them to $224 \times 224$ pixels and standardize them using mean and variance across each channel computed from the ImageNet dataset.
We use the outputs of the last layer before the final average pooling, resulting in a spatial feature map of size $512 \times 7 \times 7$.

\subsection{Feature Aggregation}
\label{subsec:aggregation}

As the memory limitations of the challenge prohibit us to store the full feature maps in memory, we choose to aggregate them into feature vectors.
This also appears sensible as the dimensionality of the full feature maps is actually larger than of the input images ($3 \times 64 \times 64$), and thus learning the latent factors from feature maps might actually be harder than from the original images.

To perform the feature aggregation, we adapt a technique introduced in the context of object retrieval, called \emph{regional maximum activations of convolutions (RMAC)} \citep{Tolias2015ParticularOR}.
In object retrieval, the goal is to find the image a target object appears on from a collection of images.
\citet{Tolias2015ParticularOR} achieve this by matching a feature vector carrying the object's ``signature'' against an RMAC feature vector for each image.
To allow matching against all the different objects that appear in an image, RMAC aggregates the signatures of objects at different scales and locations into the image feature vector. 
We assume that this property of RMAC is also useful in our case, as we need to consider different objects (e.\,g. on the  MPI3d dataset~\citep{Gondal2019OnTT}, the robotic arm and the object) to find the latent factors of variation from feature maps, but we do not know the scale and location of these object a priori.

We compute RMAC by applying max-pooling operations with different kernel sizes and strides to the feature maps (without any padding), resulting in a set of 512-dimensional feature vectors.
Concretely, we use kernel sizes $1 \times 1$, $3 \times 3$, $5 \times 5$ and $7 \times 7$ with strides $1$, $2$, $2$, $1$ respectively. 
These values were experimentally found to result in good performance.
We then $\ell 2$-normalize each of the feature vectors, sum all vectors up and apply a final $\ell 2$-normalization, resulting in the aggregated feature vector.
In contrast to \citet{Tolias2015ParticularOR}, we do not apply PCA-whitening to the feature vectors before the summation.

To reduce the computational overhead, we would like to extract and aggregate the features for each image only once before training, and store them in memory.
But because the challenge only allows to sample from the dataset (rather than selectively accessing each available image), there will be some amount of duplicates among the stored feature vectors.
To increase the amount of unique latent factor combinations available during training, we sample 1000000 images from the dataset for feature extraction, albeit the dataset having only 460800 images. 
We note that this sampling introduces an unnecessary source of randomness increasing the variance between runs. 

\subsection{VAE Training}
\label{subsec:training}

Finally, we train a standard $\beta$-VAE~\citep{Higgins2017betaVAELB} on the set of aggregated feature vectors resulting from the previous step.
The encoder network consists of three fully-connected layers with $256$, $128$, $64$ neurons, followed by two fully-connected layers parametrizing $C=18$ means and log variances of a normal distribution $\mathcal{N} \left(\vec{\mu}(\vec{x}), \vec{\sigma}^2(\vec{x}) \right)$ used as the approximate posterior $q\left(\vec{z} \mid \vec{x}\right)$.
The number of latent factors was experimentally determined.
The decoder network consists of three fully-connected layers with $64$, $128$, and $256$ neurons, followed by a fully-connected layer parametrizing the means of a normal distribution $\mathcal{N} \left(\vec{\hat{\mu}}(\vec{z}), \vec{I} \right)$ used as the conditional likelihood $p\left(\vec{x} \mid \vec{z}\right)$.
All fully-connected layers but the final ones use batch normalization and are followed by ReLU activation functions.
We use the standard Pytorch initialization for all layers and assume a factorized standard normal distribution $\mathcal{N}\left(\vec{0}, \vec{I}\right)$ as the prior $p\left(\vec{z}\right)$ on the latent variables.

For optimization, we use the Adam optimizer~\citep{Kingma2014AdamAM} with a learning rate of $0.001$, $\beta_0=0.999$, $\beta_1=0.9$ and a batch size of 256.
The VAE is trained for $N=20$ epochs by maximizing the evidence lower bound, which is equivalent to minimizing 

$$\sum_{i=1}^{512} \left(\hat{\mu}_i - x_i\right)^2 - 0.5 \frac{\beta}{C} \sum_{j=1}^C 1 + \log(\sigma_j^2) - \mu_j^2 - \sigma_j^2$$

where $\beta$ is a hyperparameter to balance the MSE reconstruction and the KLD penalty term.
As the scale of the KLD term depends on the numbers of latent factors $C$, we normalize it by $C$ such that $\beta$ can be varied independently of $C$.
It can be harmful to start training with too much weight on the KLD term~\citep{Bowman2015GeneratingSF}.
Therefore, we use the following cosine schedule to smoothly anneal $\beta$ from $\beta_\text{start}=10^{-4}$ to $\beta_\text{end}=0.12$ over the course of training:

$$\beta(t) = \begin{cases}
\beta_\text{start} &\mbox{for } t < t_\text{start} \\
\beta_\text{end} - \frac{1}{2} \left(\beta_\text{end} - \beta_\text{start} \right) \left(1 + \cos \pi \frac{t - t_\text{start}}{t_\text{end} - t_\text{start}} \right) &\mbox{for } t_\text{start} \leq t \leq t_\text{end}\\
\beta_{\text{end}} &\mbox{for } t > t_\text{end}
\end{cases}$$

where $\beta(t)$ is the value for $\beta$ in training episode $t \in \{0, \dots, N - 1\}$, and annealing runs from epoch $t_\text{start}=1$ to epoch $t_\text{end}=19$.
This schedule lets the model initially learn to reconstruct the data and only then puts pressure on the latent variables to be factorized which we found to considerably improve performance.

\section{Conclusion}

Our approach was able to obtain the second place in stage 1 of the competition.
On the public leaderboard (i.\,e. on \emph{MPI3D-realistic}), our best submission achieves the first rank on the FactorVAE \citep{Kim2018DisentanglingBF}, SAP \citep{Kumar2017VariationalIO} and DCI \citep{Eastwood2018AFF} metrics. 
See appendix \ref{apd:results} for a discussion of the results.

As \citet{Locatello2018ChallengingCA} point out, for successful unsupervised disentanglement, some kind of inductive biases are required.
We suggest that pretrained feature extractors can play the role of a strong inductive bias for natural image data. 
Our method could also be a straight-forward avenue to scale disentanglement techniques to larger image sizes.
This report only provides exploratory results, but we think that the initial results are promising enough to warrant further investigation.

\bibliography{references}

\appendix

\section{Further Notes}\label{apd:notes}

\subsection{Notes on Feature Map Extraction}

We experimented with features from pretrained ResNet, ResNeXt, DenseNet and VGG-19 architectures.
On \emph{MPI3D-simple}, ResNeXt-101 and VGG-19 outperformed ResNet and DenseNet in terms of the metrics used in the challenge. 
Between the two of them, we could not clearly detect which architecture works better.
On \emph{MPI3D-realistic} (i.\,e. on the evaluation server), VGG-19 showed better performance based on our limited number of trials, and thus we chose it as our feature extraction network.
However, we expect that ResNeXt-101 can also be used given the right kind of hyperparameter settings.

\subsection{Notes on Feature Aggregation}

Besides RMAC, we also experimented with simple spatial average- and max-pooling over the feature maps to aggregate the feature maps.
This did not result in better performance than RMAC (given the set of other hyperparameters we tested). 
We conjecture this is because global pooling loses the information of the spatial locations of objects in the image identifying some of the factors of variations.
For example, the degrees of freedom of the robotic arm can easily be derived by the relative positions of object and manipulator. 

Compared to global pooling, RMAC enhances the ability of the VAE to infer the factors of variations by better representing the properties of different objects in the image in the aggregated representation.
For example, the degrees of freedom of the robotic arm can also be derived by the specific orientation of the manipulator. However, like global pooling, RMAC also does not directly encode the spatial location of objects.
An approach to do so could be to overlay a positional encoding onto the feature maps before aggregation, for example similarly to how spatial information is encoded in self-attention mechanisms \citep{Parmar2018ImageT}.

We also experimented with PCA-whitening the regionally pooled vectors before summing them up as in the original RMAC formulation \citep{Tolias2015ParticularOR}. 
We found that this made it harder for the VAE to reconstruct the feature vectors, and thus disentanglement performance suffered.

Finally, instead of hand-designing the aggregation operation, it could also be beneficial to learn the optimal aggregation as part of the VAE training process, e.\,g. using a transformer-based approach \citep{Vaswani2017AttentionIA}.
This was not feasible within the challenge constraints as it would have required to store the full feature map in memory.

\subsection{Notes on VAE Training}
\label{apd:training}

The number of latent factors $C$ plays an important role for the performance:
if $C$ is chosen too low, the reconstruction error can not be reduced sufficiently whereas if $C$ is chosen to high, there is not enough pressure on the latent bottleneck to disentangle the latent factors; in both cases, performance suffers.
Our best model uses $C=18$.
This is considerably higher than the number of latent factors of the MPI3D dataset (i.\,e. 7), which presumably is because feature vectors encode more information (e.\,g. about textures) than raw images, and thus a larger latent bottleneck is required to reconstruct the data.

\section{Discussion of Results on the Public Leaderboard}
\label{apd:results}

\begin{table}[tbp]
\floatconts
  {tab:results}%
  {\caption{Summary of scores and ranks of our best submission on the private and public leaderboard at the end of stage 1.}}%
  {\begin{tabular}{llccccccc}
  \toprule
  & \bfseries Dataset & \bfseries FactorVAE & \bfseries DCI & \bfseries SAP & \bfseries IRS & \bfseries MIG & \\
  \midrule
Private Score & \multirow{2}{*}{MPI3d-real} & 0.792 & 0.527 & 0.166 & 0.623 & 0.292 & $\sum 2.400$ \\
Rank (of 35) & & 1 & 2 & 2 & 21 & 3 & $\diameter \, 5.8$ \\ 
  \midrule
Public Score & \multirow{2}{*}{MPI3d-realistic} & 0.848 & 0.536 & 0.183 & 0.598 & 0.347 & $\sum 2.512$ \\
Rank (of 35) & & 1 & 1 & 1 & 26 & 4 & $\diameter \, 6.6$ \\ 
  \bottomrule
  \end{tabular}}
\end{table}
We summarize the results of our best submission on the public and private leaderboards in table \ref{tab:results}. 
On the private leaderboard (i.\,e. on \emph{MPI3D-real}), our approach achieves the first rank on the FactorVAE~ \citep{Kim2018DisentanglingBF} metric, with a particularly large difference of $0.11$ to the second ranked entry.
Our submission is also second ranked on DCI~\citep{Eastwood2018AFF} and SAP~\citep{Kumar2017VariationalIO}, with small differences of respectively $0.017$ and $0.012$ to the first ranked entries.
Compared to the simulation dataset \emph{MPI3D-realistic}, there is a slight drop across all metrics besides IRS~\citep{Suter2019RobustlyDC}, reflecting the increased difficulty of disentangling natural images compared to simulation data.

On the public leaderboard (i.\,e. on \emph{MPI3D-realistic}), our method achieves the first rank on FactorVAE, SAP and DCI.
On FactorVAE, there is a particularly large margin of $0.19$ absolute difference to the second ranked method.
On MIG~\citep{Chen2018IsolatingSO}, our method achieves the fourth rank, with $0.044$ absolute difference to the best method on this metric.
Our method only falls behind on IRS, where the method is ranked 26th, with $0.145$ absolute distance to the best method.
In our experiments, there seemed to be a correlation between IRS and the amount of pressure on factorizing the latent factors (i.\,e. the $\beta$ value in the loss function). 
As a consequence, if training collapses and the KLD loss term approaches zero, the IRS can still reach high values.
This explains the number of submissions with higher IRS values (but considerably lower scores on the other metrics) than our method.
In particular, the default submission has an IRS value of $0.6199$, but fails to provide good disentanglement otherwise.
Overall, we think that the results show the potential of our approach. 

\end{document}